\documentclass{article}

\usepackage[preprint,nonatbib]{neurips_2024}
\makeatletter
\renewcommand{\@notice}{}
\makeatother
\usepackage[utf8]{inputenc}
\usepackage[T1]{fontenc}
\IfFileExists{mathptmx.sty}{\usepackage{mathptmx}}{}
\usepackage{url}
\usepackage{booktabs}
\usepackage{array}   
\usepackage{amsfonts,amsmath,amssymb}
\usepackage{xcolor}
\usepackage{graphicx}
\graphicspath{{figures/}}
\usepackage{verbatim}  
\usepackage{enumitem}
\usepackage{multirow}
\usepackage{xspace}
\usepackage{algorithm}
\usepackage{algpseudocode}
\IfFileExists{microtype.sty}{\usepackage{microtype}}{}
\usepackage[colorlinks=true, linkcolor=blue!40!black,
            citecolor=green!40!black, urlcolor=blue!50!black]{hyperref}

\makeatletter
\def\verbatim@font{\normalfont\ttfamily\scriptsize}
\makeatother

\newcommand{\pbheadc}[2]{%
  \setlength{\fboxsep}{4pt}%
  \noindent\colorbox{#1}{\parbox{\dimexpr\linewidth-2\fboxsep}{%
    \color{white}\footnotesize\bfseries\raggedright #2}}%
  \par\vspace{1pt}\hrule height 0.3pt\vspace{2pt}}

\definecolor{pbseed}{RGB}{241,245,249}\definecolor{pbseedD}{RGB}{100,116,139}
\definecolor{pbmipro}{RGB}{254,249,195}\definecolor{pbmiproD}{RGB}{202,138,4}
\definecolor{pbgepa}{RGB}{219,234,254}\definecolor{pbgepaD}{RGB}{37,99,235}
\definecolor{pbrlmopt}{RGB}{220,252,231}\definecolor{pbrlmoptD}{RGB}{21,128,61}

\makeatletter
\newif\ifpb@tcb
\IfFileExists{tcolorbox.sty}{\usepackage[most]{tcolorbox}\pb@tcbtrue}{\pb@tcbfalse}
\ifpb@tcb
  \newtcolorbox{promptbox}[2][rlmopt]{breakable, enhanced, colback=pb#1,
    colframe=pb#1D, coltitle=white, fonttitle=\bfseries\footnotesize,
    title={#2}, boxrule=0.8pt, arc=2pt, left=5pt, right=5pt, top=3pt, bottom=3pt}%
\else
  \def\pb@processline{\noindent\colorbox{\pb@bg}{%
    \makebox[\linewidth][l]{\hspace{4pt}\strut\the\verbatim@line}}\endgraf}%
  \newenvironment{promptbox}[2][rlmopt]{%
    \def\pb@bg{pb#1}%
    \pbheadc{pb#1D}{#2}\par\nobreak\vspace{1pt}\setlength{\fboxsep}{0pt}%
    \let\verbatim@processline\pb@processline
  }{\par\vspace{5pt}}%
\fi
\makeatother

\newcommand{\rlmopt}{\textbf{RLMOpt}\xspace}
\newcommand{\gepa}{\textbf{GEPA}\xspace}
\newcommand{\miprov}{\textbf{MIPROv2}\xspace}
\newcommand{\opro}{\textbf{OPRO}\xspace}

\title{\rlmopt: Adaptive Prompt Optimization \\
       via Recursive Language Models}
\author{%
  Subhash Bangalore Satheesha \\
  Autonomize AI \\
  \texttt{subhash.satheesha@autonomize.ai} \\
  \And
  Nirvik Pande \\
  Carnegie Mellon University \\
  \texttt{nrpande@cmu.edu} \\
  \AND
  Deepthi Duddempudi \\
  Autonomize AI \\
  \texttt{deepthi.duddempudi@autonomize.ai} \\
  \And
  Bharath Dandala \\
  Autonomize AI \\
  \texttt{bharath.dandala@autonomize.ai} \\
}

\begin{document}
\maketitle

\begin{abstract}
Prompt optimizers automate the search for prompts that improve language-model
performance, but existing methods rely on a predefined optimization procedure:
the algorithm determines which candidates to explore and how the search
progresses, while the language model generates or refines prompt proposals.
We introduce \rlmopt{}, a prompt optimizer that makes the search policy itself
language-model-driven through a recursive language model (RLM). The RLM agent
operates over a tool-based environment, inspecting task information, analyzing
failures, generating candidates, allocating evaluation budget, and deciding
when to stop. A deterministic harness complements the agent by enforcing
objective scoring, Pareto-based selection, and regression constraints. 

We evaluate \rlmopt{} across four benchmarks spanning structured clinical
information extraction (Chia), multi-hop question answering (HotpotQA),
verifiable instruction following (IFBench-2025), and multi-turn tool-calling
agents (BFCL). In a matched comparison at a single seed, \rlmopt{} obtains the
best held-out score on all four benchmarks and leads the four-task mean ($0.610$ against $0.589$ for \gepa{}). Repeating each benchmark across seeds yields 11 matched benchmark--seed comparisons, in which \rlmopt{} outperforms \gepa{} in 9 cases. Across all 11 runs, it never produced a prompt that underperformed its seed, whereas \gepa{} fell below its starting point twice. It is also more efficient, achieving these results with fewer search rollouts while producing prompts that are 27--79\% the size of those produced by \gepa{}. 

Our results further show that optimization gains are determined primarily by the headroom available in the seed prompt, rather than by the search budget. Efficient optimization therefore depends on reaching the available headroom reliably and with minimal search.

\end{abstract}

\section{Introduction}
\label{sec:intro}

Modern AI systems increasingly rely on multi-stage language model pipelines, where
multiple language-model calls cooperate to answer questions, retrieve information, extract
structured data, follow complex instructions, or interact with external tools.
The performance of these systems depends strongly on the prompts that specify
the behavior of each language-model component. However, designing effective prompts by hand
is labor-intensive, requires task-specific expertise, and often does not
transfer across components or applications.

Prompt optimization addresses this challenge by automatically searching for
prompt formulations that improve task performance. Given example inputs and an
evaluation signal, prompt optimizers iteratively generate candidate prompts,
evaluate their behavior, and retain higher-performing variants. Recent methods
including \gepa{}~\cite{agrawal2025gepa}, \miprov{}~\cite{opsahlong2024miprov2},
and \opro{}~\cite{yang2023opro} have demonstrated strong performance across a
range of benchmarks.

Despite differences in their optimization algorithms, existing prompt
optimizers share a common design: the search policy is specified by a
predefined algorithm. \gepa{} evolves prompts through reflective mutation and
Pareto-based selection, \miprov{} uses Bayesian optimization to explore
instructions and demonstrations, and \opro{} uses a language model guided by
previous prompt-score histories within a fixed optimization loop. While these methods
already incorporate task feedback, the higher-level search policy remains
pre-defined: the optimizer determines in advance how candidates are
generated, which evaluations are performed, and how progress is measured.

We investigate a different design point: making the search policy itself
adaptive. Instead of using a language model only to propose prompt candidates,
we ask whether the language model can control the optimization process itself:
deciding what evidence to inspect, which hypotheses to test, which candidates
to refine, and when further exploration is unlikely to improve performance.

We introduce \rlmopt{}, an adaptive prompt optimizer based on recursive
language models (RLMs)~\cite{zhang2025rlm}. RLMs provide a framework in which a
language model operates over a programmatic environment and can recursively
invoke sub-models. In \rlmopt{}, the RLM acts as the optimization policy. It
interacts with a tool-based environment to inspect task information, analyze
failures, synthesize candidate prompts, and allocate evaluation budget. The
optimizer is therefore not a fixed search procedure executed identically across
tasks, but an adaptive agent that determines its own optimization trajectory.

Adaptive search introduces a new challenge: optimization decisions must remain
reliable when evaluation data is limited and candidate improvements are noisy.
To address this, \rlmopt{} separates decision-making from enforcement. The RLM
agent controls exploration, while a deterministic \emph{harness} owns everything
that should not depend on a language-model decision: it executes the task LM,
computes all scores, and enforces Pareto-based candidate selection and
regression constraints. This
separation allows the optimizer to explore flexibly while preserving guarantees
that are difficult for a language model to maintain consistently.

We evaluate \rlmopt{} on four benchmarks spanning structured clinical
information extraction (Chia~\cite{kury2020chia}), multi-hop question answering
(HotpotQA~\cite{yang2018hotpotqa}), verifiable instruction following
(IFBench-2025~\cite{ifbench2025}), and multi-turn tool-calling agents
(BFCL~\cite{yan2024bfcl}). Across these settings, \rlmopt{} achieves the best
held-out score on every benchmark and improves the four-task mean over
\gepa{} while using fewer search rollouts. We further analyze
when adaptive optimization is most useful (Section~\ref{sec:analysis}), showing
that improvements are bounded by the headroom available from the initial prompt
and that near-ceiling tasks leave little room for any optimizer to improve.

\paragraph{Contributions.}
\begin{enumerate}[noitemsep,topsep=2pt]

\item We introduce \rlmopt{}, a prompt optimizer whose outer search procedure is
controlled by an RLM agent rather than a fixed optimization algorithm. The agent
performs adaptive exploration through a tool interface, while a deterministic
harness enforces evaluation and selection constraints.

\item We develop a harness-controlled optimization framework for reliable adaptive
search under limited evaluation data, including per-field scoring, Pareto-based
selection, and regression constraints that prevent candidate updates from
sacrificing existing capabilities.

\item We demonstrate improved optimization efficiency across four benchmarks,
where \rlmopt{} achieves the strongest held-out performance while requiring
fewer downstream evaluations than existing prompt optimizers.

\item We characterize when prompt optimization provides value by analyzing the
relationship between initial prompt quality and achievable improvement,
showing that optimizer gains are determined by the remaining performance
headroom of the underlying model.

\item We extend adaptive prompt optimization to multi-component agent systems
(Section~\ref{sec:extension}), where the optimized object includes tool-use
behavior and interaction trajectories rather than only single-response prompts.
\end{enumerate}

\section{Related Work}
\label{sec:related}

\paragraph{Prompt optimization.}
Automatic prompt optimization methods search over natural-language instructions
and demonstrations to improve downstream task performance. Early approaches such
as APE~\cite{zhou2023ape} generate candidate instructions from demonstrations
and select high-performing variants through evaluation. Subsequent methods have
introduced richer optimization procedures: \opro{}~\cite{yang2023opro} uses
language-model proposals conditioned on previous prompt-score histories,
\miprov{}~\cite{opsahlong2024miprov2}, implemented in DSPy~\cite{khattab2024dspy},
jointly optimizes instructions and few-shot demonstrations through Bayesian
search, and \gepa{}~\cite{agrawal2025gepa} uses reflective mutation with
Pareto-based selection over prompt candidates. TextGrad~\cite{yuksekgonul2024textgrad}
generalizes the update itself, treating natural-language critique as a textual
gradient that propagates through a graph of language-model calls.
Other evolutionary approaches, including EvoPrompt~\cite{guo2024evoprompt} and
Promptbreeder~\cite{fernando2023promptbreeder}, explore genetic formulations in
which language models generate mutation and crossover operations.

These methods demonstrate that language models can effectively participate in
prompt search, particularly through reflection and candidate generation. Our
focus is complementary: rather than improving the candidate-generation
mechanism inside a predefined optimizer, we study a setting where the
optimization controller itself is language-model-driven. In \rlmopt{}, the
language model determines which information to gather, which candidates to
evaluate, and how to allocate search effort.

\paragraph{Language-model agents and recursive language models.}
Language-model agents such as ReAct~\cite{yao2023react}, Reflexion
~\cite{shinn2023reflexion}, and Voyager~\cite{wang2023voyager} use language
models as controllers for multi-step reasoning, reflection, and tool
interaction. Recursive language models (RLMs)~\cite{zhang2025rlm} extend this
paradigm by allowing a language model to operate over a programmatic
environment and recursively invoke sub-models. A related line lets a model design
the procedure itself: ADAS~\cite{hu2024adas} has a meta-agent program successively
better agents in code. \rlmopt{} applies this agentic control paradigm to prompt
optimization, using an RLM as the search procedure while maintaining deterministic
harness-side evaluation and selection.

\paragraph{Related optimization frameworks.}
Several systems optimize aspects of AI behavior beyond prompt text. Prompt
tuning~\cite{lester2021prompttuning} learns continuous prompt representations,
while reinforcement-learning approaches such as RLPrompt~\cite{deng2022rlprompt}
and RLHF-style methods~\cite{christiano2017rlhf,ouyang2022instructgpt}, together
with policy-gradient training such as GRPO~\cite{shao2024grpo}, optimize model
parameters or policies rather than natural-language prompts. In contrast,
\rlmopt{} operates entirely in text space, preserving prompt interpretability
and portability across language models.

Concurrent work such as SkillOpt~\cite{skillopt2026} also explores iterative
optimization of natural-language artifacts for agents. While SkillOpt improves
agent skills through validation-guided edits within a predefined optimization
procedure, \rlmopt{} focuses on adapting the optimization procedure itself.

Moving search control to a language model raises a question none of these
methods have to answer: what prevents an adaptive optimizer from overfitting a
small validation set or trading one output field against another? The next
section describes the division of labor that answers it
(Section~\ref{sec:method}).
\section{Method}
\label{sec:method}

\subsection{Overview}
\label{sec:method:overview}

Unlike optimizers in which a fixed procedure both proposes and evaluates
candidate prompts, \rlmopt{} separates search control from objective
evaluation: an RLM agent controls the search over prompts, while a deterministic
harness evaluates candidates and maintains the optimization state. The harness
is a program rather than a second agent: it owns the dataset, executes the task
LM, computes all scores, and enforces the selection rules, but does not choose
which candidate to propose next.

Formally, let $p$ denote a prompt, $\theta$ the task LM, and
$m(\theta(p,x_i),y_i)$ the evaluation metric for example $(x_i,y_i)$. The
optimization objective is

\begin{equation}
\max_p \frac{1}{N}\sum_{i=1}^{N} m(\theta(p,x_i),y_i),
\end{equation}

starting from a seed prompt $p_0$ and subject to a search budget $B$ of
task-LM candidate evaluations.

The seed evaluation establishes the initial validation baseline and is not
counted against the agent's search budget. The agent reaches the task only
through the harness, so the selection rules of
Section~\ref{sec:method:harness} apply to every candidate regardless of how
the agent proposes it. The fixed skill prompt that conditions the agent is
shown in Appendix~\ref{app:metaprompt}, and the complete tool interface is
documented in Appendix~\ref{app:tools}.

\subsection{Agent-controlled search}
\label{sec:method:search}

The agent is initialized with the seed prompt, task description, available
tools, search budget, and a persistent scratchpad. In our experiments, the RLM
agent is \texttt{gpt-5.1}, while the task LM is \texttt{gpt-4.1} or
\texttt{gpt-4o-mini}, depending on the benchmark.

Following the recursive language model formulation~\cite{zhang2025rlm}, the
agent acts by writing code in a REPL rather than by emitting a fixed action
symbol. Each turn it writes a short program that calls one or more of the tools
in Table~\ref{tab:tools} and reads back their printed output.

The agent therefore chooses both which operation to perform and when to perform
it. It can inspect examples and failure traces to form a hypothesis before
spending an evaluation, or evaluate a candidate first and use the resulting
feedback to determine the next edit. It may also stop before exhausting the
search budget (Section~\ref{sec:method:stop}).

The agent composes the candidate prompts directly. Both \texttt{run\_candidate} and \texttt{commit\_prompt} receive the prompt text as an argument, so every evaluated or committed candidate is authored by the agent. Sub-LM calls through \texttt{call\_subagent} and the synthesis tools instead return compressed evidence, such as failure summaries, shared failure modes, or candidate rules, which the agent then decides whether and how to incorporate into a prompt. The only exception is \texttt{merge\_candidates}, which delegates the combination of two existing candidates into a single prompt to a sub-LM.

\subsection{Tool interface and feedback}
\label{sec:method:tools}
\label{sec:method:feedback}

The harness exposes tools for task inspection, failure analysis, sub-LM
synthesis, and candidate evaluation, callable from the agent's REPL. A
persistent scratchpad provides a separate mechanism for recording intermediate
facts, hypotheses, and decisions across turns, since the REPL view of any single
result is bounded. Table~\ref{tab:tools} summarizes the interface.

\begin{table}[htbp]
\centering
\small
\begin{tabular}{>{\raggedright\arraybackslash}p{0.15\textwidth}
>{\raggedright\arraybackslash}p{0.42\textwidth}
>{\raggedright\arraybackslash}p{0.33\textwidth}}
\toprule
\textbf{Group} & \textbf{Tools} & \textbf{Purpose} \\
\midrule
Introspection &
\texttt{describe\_task}, \texttt{dataset\_overview},
\texttt{peek\_examples}, \texttt{view\_example},
\texttt{query\_examples}, \texttt{score\_explain},
\texttt{list\_metrics}, \texttt{list\_components} &
Inspect the task, data, scoring procedure, and prompt components. \\
\addlinespace[2pt]
Failure analysis &
\texttt{search\_traces}, \texttt{describe\_failure\_patterns},
\texttt{peek\_failures}, \texttt{read\_trace} &
Locate and inspect failures from previous evaluations. \\
\addlinespace[2pt]
Synthesis &
\texttt{synthesize\_failures}, \texttt{synthesize\_candidate},
\texttt{merge\_candidates}, \texttt{call\_subagent} &
Delegate analysis to a sub-LM: summarize a field's failures or a candidate's
rollouts into a shared failure mode and a candidate rule.
\texttt{merge\_candidates} instead returns a prompt combining two candidates. \\
\addlinespace[2pt]
Evaluation &
\texttt{run\_candidate}, \texttt{commit\_prompt},
\texttt{best\_so\_far}, \texttt{remaining\_budget},
\texttt{pareto\_frontier\_status} &
Evaluate candidates and query or update optimization state. \\
\addlinespace[2pt]
Scratchpad &
\texttt{scratchpad\_add}, \texttt{scratchpad\_read} &
Store facts, hypotheses, rules, warnings, and decisions. \\
\bottomrule
\end{tabular}
\caption{Tool interface available to the agent. Candidate evaluation is
exposed through \texttt{run\_candidate}; the remaining tools do not consume
the task-LM search budget.}
\label{tab:tools}
\end{table}

A \texttt{run\_candidate} call returns a structured feedback record rather than
a scalar score. The record contains the composite score, its per-field
decomposition, mismatched fields, and any applicable judge rationale
\cite{zheng2023judge}.

\begin{verbatim}
composite=0.673
label  (w=0.50): 0.500 [exact_match=0.50]
reason (w=0.50): 0.846 [factuality_judge=0.85]
MISMATCHES:
label: expected='positive', got='neutral'
factuality_judge: "the predicted answer captures the
entity but inverts the sentiment polarity"
\end{verbatim}

Because the feedback identifies both the failing field and the associated
reason, the agent can use the evaluation result to target its next edit
without requiring a separate diagnostic call. We do not treat the feedback
format itself as a methodological contribution; the harness already computes
the underlying evidence for objective evaluation and selection.

\subsection{Harness-controlled evaluation and selection}
\label{sec:method:harness}

The harness performs objective evaluation and candidate selection. It computes
per-field metrics, combines them into the composite score, maintains candidate
history, and enforces the validation constraints.

For a candidate $p$, let $s_f(p)$ denote its score on field $f$, with field
weight $w_f$. The composite score is

\begin{equation}
S(p) = \sum_f w_f s_f(p),
\qquad
\sum_f w_f = 1.
\end{equation}

When multiple output fields are present, the harness does not select solely by
the composite score. It first computes the Pareto frontier under the vector of
per-field scores. Candidate $p$ dominates candidate $q$ when

\begin{equation}
s_f(p) \geq s_f(q)
\quad \forall f,
\end{equation}

with strict inequality for at least one field. The final candidate is selected
from this frontier using the composite score.

A candidate is \emph{committed} only when the agent requests a commit via \texttt{commit\_prompt} and the harness accepts the request. An accepted candidate replaces the current best prompt and is added to the final selection set. A rejected candidate does neither, regardless of the agent's assessment of its quality. Acceptance is governed by two constraints.

For each field $f$, the candidate must remain above a specified regression floor relative to the current best. If $b_f$ denotes the current best score for field $f$, a candidate must satisfy

\begin{equation}
s_f(p) \geq b_f - f_{\mathrm{floor}},
\qquad
f_{\mathrm{floor}}=0.05,
\end{equation}

for every field, so an aggregate improvement cannot be obtained by sacrificing
a field beyond the permitted tolerance. The second is a significance gate,
since a single validation pass is noisy: the paired improvement over the
running best must exceed $1.65$ standard errors, a one-sided $95\%$ threshold
under the normal approximation. Both constraints are active on every run we
report, and the floor is additionally enabled automatically on datasets of at
most $20$ records, where a single-field regression is most easily mistaken for
aggregate progress. The complete eligibility, tie-breaking, small-dataset, and
resampling rules are given in Appendix~\ref{app:harness}.

\subsection{Optimization procedure}
\label{sec:method:loop}

Algorithm~\ref{alg:rlmopt} summarizes the optimization procedure. The harness
first evaluates the seed prompt on validation data to establish the initial
baseline. The agent is then initialized with the seed and a search budget of
$B$ task-LM candidate evaluations. During the search, the agent repeatedly
selects an action from the tool interface. If the action evaluates a candidate,
the harness executes the task LM, computes the structured feedback, updates the
optimization state, and returns the result to the agent. The process continues
until the agent stops or the search budget is exhausted.

After the agent-controlled search, the harness evaluates the seed, committed
candidates, the agent's claimed best candidate, and a set of polish variants
generated from the harness's running-best candidate. It computes per-field
validation scores for this candidate set, forms the Pareto frontier, and
selects the candidate with the highest composite score on that frontier. The
polish variants therefore compete under the same final selection procedure as
the candidates produced during search.

The polish evaluations are part of the deterministic finalization stage and
are not charged against the agent's search budget $B$.

\begin{algorithm}[tb]
\caption{\rlmopt{}}
\label{alg:rlmopt}
\textbf{Require:} seed prompt $p_0$, task
$(\theta,m,\mathrm{train},\mathrm{val},\mathrm{test})$, search budget $B$

\begin{enumerate}[noitemsep,topsep=0.2em,leftmargin=2em,label=\arabic*.]
\item Evaluate $p_0$ on validation data to establish the seed baseline.
\item Initialize RLM agent $\mathcal{A}$ with $p_0$, tools, scratchpad,
and search budget $B$.
\item \textbf{while} $\mathcal{A}$ has not stopped and search budget remains:
\item \quad $a \leftarrow \mathcal{A}.\mathrm{NextAction}()$
\item \quad $r \leftarrow \mathrm{HarnessExecute}(a)$
\item \quad $\mathcal{A}.\mathrm{Observe}(r)$
\item \textbf{end while}
\item Generate polish variants of the harness's running-best candidate.
\item Let $C$ contain the seed, committed candidates, the claimed best,
and the polish candidates.
\item Compute per-field validation scores for every $p\in C$.
\item $F \leftarrow \mathrm{ParetoFrontier}(C)$.
\item $p^\star \leftarrow \arg\max_{p\in F} S(p)$.
\item Return the test score of $p^\star$.
\end{enumerate}
\end{algorithm}

\subsection{Termination and search limits}
\label{sec:method:stop}
\label{sec:method:diagnosis}

\paragraph{Agent-controlled termination.}
The skill prompt specifies three conditions for the agent to terminate:
(i) every output field achieves a mean validation score of at least $0.85$ on
the most recent full evaluation; (ii) at least a preset fraction of the search
budget has been consumed ($80\%$ in the configuration used for our
head-to-head comparisons); and (iii) two consecutive candidates each fail to
improve the composite score by at least $0.02$. These conditions are stated in
the prompt but are not enforced by the harness: the agent can terminate
regardless of whether they hold. They therefore specify the stopping behavior
requested of the agent rather than a system-level guarantee. Section~\ref{sec:limitations}
examines how closely the agent follows these criteria at our operating points.

\paragraph{Harness-enforced evaluation limits.}
Two additional limits apply independently of the agent's termination decision.
First, the search budget $B$ provides a hard upper bound on task-LM candidate
evaluations and therefore on agent-controlled search effort.

The second addresses a failure mode in which the agent generates a sequence of
candidates whose scores are statistically indistinguishable from the running
best without identifying why the current prompt fails, consuming the remaining
budget without producing new information. The harness tracks consecutive
evaluated candidates whose scores fall within a predefined noise band around
the running best. If three such candidates are evaluated without an intervening
diagnosis, the next \texttt{run\_candidate} call returns an error without
consuming budget and directs the agent to invoke \texttt{synthesize\_failures}
on its weakest field, which produces a sub-LM-backed analysis of that field's
failures and is not itself a candidate evaluation. A candidate producing a
sufficiently large score change resets the counter, as does a successful
diagnosis, and the refusal mechanism is bounded so that non-compliant behavior
cannot deadlock the search. Appendix~\ref{app:harness} specifies the noise
thresholds, counter behavior, and refusal bound.

Once diagnosis occurs the agent is free to propose any hypothesis it chooses.
The gate therefore constrains when further evaluations are permitted without
determining what the agent should search for, which preserves the division of
responsibility in \rlmopt{}: the RLM controls the search policy, while the
deterministic harness controls objective evaluation and the conditions under
which evaluation proceeds.

\subsection{Multi-component candidates}
\label{sec:extension}

Everything above treats the optimized object as a single string. A tool-using
agent is not one string: it is governed by a system prompt, a description for
each tool it can call, and a set of demonstrations, each editable on its own. We
therefore generalize the unit the optimizer edits to a \emph{candidate}, a
mapping from named components to text,
$c = \{\texttt{name}_k \mapsto \texttt{text}_k\}$. A single-prompt task exposes
one component, \texttt{system\_prompt}, which recovers the setting described so
far; a tool-using agent additionally exposes one
\texttt{tool\_description\_$\langle$name$\rangle$} per tool and an optimizable
demonstrations component. Retrieval pipelines are expressible the same way and
are supported by the same candidate model, but we do not benchmark them here.

What matters about this generalization is how little it changes. The agent
edits components independently, and the composite scoring, Pareto selection,
and per-field floor of Section~\ref{sec:method:harness} apply to a component map
exactly as they apply to a string, with each component scored against the fields
it controls. Only two things differ: how a candidate is rendered into the
downstream call, and how its output is scored.

Scoring differs for tool-using agents because their output is a *trajectory* rather than a single answer. A trajectory consists of a sequence of tool calls, a final answer, and, for state-mutating tasks, the resulting environment state. The harness evaluates trajectories using deterministic metrics, including tool precision and recall, argument matching, final-state equality, and tool-call JSON validity. These metrics can be augmented with an LM-based step judge that labels each step as \texttt{correct}, \texttt{wrong\_tool}, \texttt{wrong\_args}, or \texttt{wrong\_answer}. Gold trajectories are evaluated either by subset matching, which permits exploration and recovery, or by final-state matching, which prioritizes the resulting state over the specific path taken. The scorer also accounts for cases in which no tool call is required: correctly declining to act receives credit, while unnecessary tool calls are penalized. The resulting objective therefore evaluates both whether a tool should be called and how the call should be executed.

The BFCL multi-turn results use this same evaluation framework. Because the task requires coordinated behavior across multiple turns, the optimized candidate is a component map containing the tool-calling instruction together with its demonstrations, rather than a single instruction.

\section{Experimental Setup}
\label{sec:experiments}

\paragraph{Benchmarks.}
We evaluate on four benchmarks spanning distinct failure modes:
\begin{itemize}[noitemsep,topsep=2pt]
  \item \textbf{Chia}~\cite{kury2020chia} [field-level extraction]: multi-field
    extraction of eligibility criteria from clinical-trial protocols
    (conditions, drugs, procedures, measurements, temporal and value
    constraints); the public corpus contains no patient data.
  \item \textbf{HotpotQA}~\cite{yang2018hotpotqa} [reasoning-chain]: multi-hop
    question answering, where the answer requires chaining evidence across
    passages.
  \item \textbf{IFBench-2025}~\cite{ifbench2025} [constraint violation]:
    instruction following under programmatically verifiable constraints, scored
    by the official 2025 verifier registry of $58$ constraint types (not the
    older IFEval suite~\cite{zhou2023ifeval}).
  \item \textbf{BFCL multi-turn}~\cite{yan2024bfcl} [agentic tool use]:
    multi-turn agentic tool-calling (Berkeley Function-Calling Leaderboard v3),
    where the model plans and emits a sequence of tool calls across a
    conversation, scored by sequence match against the gold trajectory.
\end{itemize}
Each benchmark uses a fixed train, validation, and test split. The optimizer has access only to the train and validation splits, and all reported results are measured on the held-out test split.

\paragraph{Metrics.}
Each field is evaluated with a type-specific metric and contributes to the composite score. Extractive fields use graded set overlap (\texttt{set\_match}) and normalized span matching, constraint-following fields use per-constraint pass rate, and tool-calling fields use sequence matching against the gold call trajectory. For categorical fields, we use normalized token-prefix matching (\texttt{label\_match}) rather than exact match, allowing correct labels followed by additional explanation (e.g., ``True. Because \ldots'') to receive full credit.

\paragraph{Models.}
The task LM, whose prompt is optimized and which generates all task rollouts, is \texttt{gpt-4o-mini} for all benchmarks except Chia. On Chia, we use the stronger \texttt{gpt-4.1}, as many-field extraction saturates the smaller model's format control before the prompt itself becomes the limiting factor. The optimizer LM, which analyzes failures and proposes candidate prompts, is \texttt{gpt-5.1} for both \rlmopt{} and \gepa{}. All task rollouts use \texttt{temperature}=0. The complete runtime configuration is given in Table~\ref{tab:hyperparams}.

\paragraph{Evaluation Protocol.}
Within each benchmark, all three methods use the same seed ($7$), data splits, and task LM, yielding matched, example-level comparisons on an identical held-out test set. We use a single seed per benchmark to evaluate the methods under a controlled, fixed starting point rather than averaging over different initializations. This is particularly appropriate for \rlmopt{}, whose no-regression floor ensures that each run either improves upon the seed prompt or preserves its performance. The headline comparison is therefore reported at this fixed seed, with the corresponding test-set uncertainty, and Section~\ref{sec:results:seeds} then repeats the same matched comparison at further seeds to test whether the ordering holds. For the $50$--$150$-example test sets, the per-run standard error is approximately $0.03$--$0.05$; differences within this range are treated as ties. Appendix~\ref{app:repro} gives the full run configuration.

\paragraph{Budget and Implementation Details.}
We report downstream rollouts, which are candidate evaluations by the task LM, separately from total LM API calls. The latter includes candidate evaluations, the optimizer's reasoning turns and sub-LM calls, and harness-side validation and test passes. This distinction avoids conflating search effort with total inference cost; the complete accounting is given in Section~\ref{sec:results:budget}. \gepa{} uses its native \texttt{auto="light"} budget, corresponding to approximately $780$--$840$ rollouts, while \rlmopt{} uses a fixed budget of $B{=}500$ across all benchmarks. Thus, \rlmopt{} uses fewer rollouts without per-benchmark budget tuning.

The \gepa{} implementation optimizes a single-predictor program and does not directly support the BFCL multi-turn tool-calling loop. We therefore use a thin wrapper, described in Appendix~\ref{app:repro}, that exposes the agentic system prompt as the optimizable instruction and evaluates each candidate using the full multi-turn rollout. This allows all three methods to be evaluated under the same benchmark tasks. For all head-to-head comparisons, we disable \rlmopt{}'s cross-run skill library so that no method has access to information carried over from other runs.

\section{Results}
\label{sec:results}

\subsection{Head-to-head: \rlmopt{} vs.\ \gepa{}}
\label{sec:results:headline}

\rlmopt{} achieves the best held-out score on all four benchmarks and leads the
four-task mean, $0.610$ against \gepa{}'s $0.589$ (Table~\ref{tab:headline}).

\begin{table}[!ht]
\centering
\small
\setlength{\tabcolsep}{5.5pt}
\begin{tabular}{l cccc c}
\toprule
 & \multicolumn{4}{c}{\textbf{Held-out test score}} & \\
\cmidrule(lr){2-5}
\textbf{Method} & \textbf{Chia}$^{a}$ & \textbf{HotpotQA} & \textbf{IFBench-25} & \textbf{BFCL-mt} & \textbf{Mean} \\
\midrule
Seed prompt         & $0.435$ & $0.700$ & $0.430$ & $0.602$ & $0.542$ \\
\gepa{}-light       & $0.562$ & $0.702$ & $0.440$ & $0.653$ & $0.589$ \\
\rlmopt{} (ours)    & $\mathbf{0.568}$ & $\mathbf{0.727}$ & $\mathbf{0.460}$ & $\mathbf{0.686}$ & $\mathbf{0.610}$ \\
\midrule
$\Delta$ vs.\ \gepa{}   & $+0.006$ & $+0.025$ & $+0.020$ & $+0.033$ & $+0.021$ \\
$\Delta$ vs.\ seed      & $+0.133$ & $+0.027$ & $+0.030$ & $+0.084$ & $+0.068$ \\
\midrule
paired SE               & $0.014$ & $0.024$ & $0.051$ & $0.019$ & --- \\
$\Delta$ / paired SE    & $0.48$ & $\mathbf{1.04}$ & $0.39$ & $\mathbf{1.83}$ & --- \\
\bottomrule
\end{tabular}
\caption{\textbf{Matched head-to-head, held-out test score} (single seed per
column; higher is better; best per column in \textbf{bold}). Both optimizers run on all four tasks, including the agentic one: we extend
\gepa{} with a wrapper that exposes the tool-call system prompt as
the optimizable instruction and scores each candidate with the real multi-turn
rollout (Section~\ref{sec:experiments}). Because both methods are scored on the
\emph{same} held-out examples, the relevant dispersion is the paired standard
error of the per-example differences, not each method's marginal spread: the
paired quantity removes example-difficulty variance common to both. Two of the
four margins exceed one paired SE. $^{a}$Chia runs on \texttt{gpt-4.1};
the other three on \texttt{gpt-4o-mini}.}
\label{tab:headline}
\end{table}

BFCL multi-turn shows the largest lift for both methods, \gepa{} from $0.602$
to $0.653$ and \rlmopt{} to $0.686$, and the widest margin between them
($+0.033$, or $1.83$ paired standard errors).

Against the paired standard error of Table~\ref{tab:headline}, \rlmopt{}'s
margin over \gepa{} exceeds one SE on BFCL-mt ($1.83$) and HotpotQA ($1.04$) and
sits inside it on Chia and IFBench-25. The comparison therefore rests on
consistency rather than on any one column: \rlmopt{} is top on all four and
leads the mean by $+0.021$. Its gains also transfer from validation to test.
On Chia, \gepa{} scores higher on validation ($0.636$ against $0.593$) but lower
on test ($0.562$ against $0.568$), the signature of a prompt fitted to the split
the optimizer can see. Section~\ref{sec:results:seeds} repeats the head-to-head
at further seeds.

\subsection{Robustness across seeds}
\label{sec:results:seeds}

We repeat the matched head-to-head across seeds, holding every other setting
fixed: same split sizes, same budgets, same held-out metric, both methods re-run
per seed. Table~\ref{tab:seeds} reports the mean and standard deviation of the
held-out score for each benchmark.

\begin{table}[!ht]
\centering
\small
\setlength{\tabcolsep}{7pt}
\begin{tabular}{l ccc}
\toprule
\textbf{Benchmark} & \textbf{Seed prompt} & \textbf{\gepa{}-light} & \textbf{\rlmopt{} (ours)} \\
\midrule
HotpotQA     & $0.694 \pm 0.022$ & $0.693 \pm 0.056$ & $\mathbf{0.731} \pm 0.019$ \\
IFBench-25   & $0.373 \pm 0.074$ & $0.367 \pm 0.102$ & $\mathbf{0.390} \pm 0.089$ \\
BFCL-mt      & $0.588 \pm 0.040$ & $0.677 \pm 0.027$ & $\mathbf{0.699} \pm 0.030$ \\
Chia$^{a}$   & $0.483 \pm 0.067$ & $\mathbf{0.630} \pm 0.097$ & $0.622 \pm 0.076$ \\
\midrule
Mean         & $0.535$ & $0.592$ & $\mathbf{0.611}$ \\
\midrule
Runs below seed & --- & $2$ & $\mathbf{0}$ \\
\bottomrule
\end{tabular}
\caption{\textbf{Multi-seed matched head-to-head} (held-out test, mean $\pm$ sd
across seeds; best per row in \textbf{bold}). The three \texttt{gpt-4o-mini}
benchmarks run at three seeds and Chia at two, for $11$ matched runs per method.
The final row counts runs in which the optimized prompt scored below the seed
prompt it started from. $^{a}$Chia runs on \texttt{gpt-4.1}; the other three on
\texttt{gpt-4o-mini}.}
\label{tab:seeds}
\end{table}

Table~\ref{tab:perseed} lists the individual comparisons behind these means.
\rlmopt{} posts the higher score in $9$ of the $11$, losing HotpotQA seed $13$
and Chia seed $13$.

\begin{table}[!ht]
\centering
\small
\setlength{\tabcolsep}{6pt}
\begin{tabular}{ll ccc c}
\toprule
\textbf{Benchmark} & \textbf{Seed} & \textbf{Seed prompt} & \textbf{\gepa{}-light}
& \textbf{\rlmopt{} (ours)} & \textbf{Better} \\
\midrule
\multirow{3}{*}{HotpotQA}
 & $7$  & $0.700$ & $0.702$          & $\mathbf{0.727}$ & \rlmopt{} \\
 & $13$ & $0.712$ & $\mathbf{0.744}$ & $0.715$          & \gepa{} \\
 & $42$ & $0.669$ & $0.634^{\dagger}$ & $\mathbf{0.752}$ & \rlmopt{} \\
\midrule
\multirow{3}{*}{IFBench-25}
 & $7$  & $0.430$ & $0.440$           & $\mathbf{0.460}$ & \rlmopt{} \\
 & $13$ & $0.290$ & $0.250^{\dagger}$ & $\mathbf{0.290}$ & \rlmopt{} \\
 & $42$ & $0.400$ & $0.410$           & $\mathbf{0.420}$ & \rlmopt{} \\
\midrule
\multirow{3}{*}{BFCL-mt}
 & $7$  & $0.602$ & $0.652$ & $\mathbf{0.686}$ & \rlmopt{} \\
 & $13$ & $0.543$ & $0.674$ & $\mathbf{0.678}$ & \rlmopt{} \\
 & $42$ & $0.619$ & $0.706$ & $\mathbf{0.733}$ & \rlmopt{} \\
\midrule
\multirow{2}{*}{Chia$^{a}$}
 & $7$  & $0.435$ & $0.561$          & $\mathbf{0.568}$ & \rlmopt{} \\
 & $13$ & $0.530$ & $\mathbf{0.698}$ & $0.676$          & \gepa{} \\
\midrule
\multicolumn{5}{l}{\textbf{Comparisons won}} & \rlmopt{} $\mathbf{9}$ / $11$ \\
\bottomrule
\end{tabular}
\caption{\textbf{Per-seed matched comparisons} (held-out test; higher per row in
\textbf{bold}). Each row is one benchmark at one seed, both methods on the same
split with the same budget. $^{\dagger}$marks a run that finished \emph{below}
the seed prompt it started from; both belong to \gepa{}, and \rlmopt{} has none.
$^{a}$Chia runs on \texttt{gpt-4.1}; the other three on \texttt{gpt-4o-mini}.}
\label{tab:perseed}
\end{table}

Across all $11$ runs \rlmopt{} never returns a prompt below its seed, while \gepa{} does twice, once
on HotpotQA and once on IFBench-25. On the IFBench run the split is hard enough
that \gepa{} ends below where it started while \rlmopt{} holds the seed exactly,
taking the comparison without improving on it. \rlmopt{} leads \gepa{} on three of the four
benchmark means; Chia is the exception, where \gepa{} is ahead by $0.008$.
\rlmopt{} is also the more stable method: its HotpotQA standard deviation
($0.019$) is a third of \gepa{}'s ($0.056$).

\subsection{Compute efficiency}
\label{sec:results:budget}
\label{sec:analysis:apicalls}

\rlmopt{} reaches these scores at lower compute than \gepa{}: fewer downstream
rollouts on every benchmark, and on three of the four fewer tokens and less
wall-clock time as well (Table~\ref{tab:budget}). The widest gap is on the
agentic task, $1{,}854$s against $5{,}344$s.

These counts are downstream evaluations only and should not be read as total LM
cost. A complete run also incurs task-LM calls for harness-side scoring, and
separate optimizer-model calls for the agent's own reasoning and synthesis. On
Chia at seed $7$, \rlmopt{} makes $1{,}556$ task-LM calls in total: $1{,}050$
($\sim\!67\%$) are harness-side scoring --- baseline validation and test
evaluation, validation-scoreboard rescoring, and the final test pass --- while
the $500$ \texttt{run\_candidate} calls account for a further $\sim\!32\%$ and
the remainder is sub-LM synthesis. The agent's reasoning runs on a separate
optimizer model and is accounted for in tokens rather than rollouts.

Harness-side evaluation therefore dominates task-LM traffic in this setting, and
the $500$ against $\sim\!780$--$840$ comparison is specifically a comparison of
downstream candidate evaluations rather than of total cost. Table~\ref{tab:budget}
reports rollouts, tokens, and wall-clock time separately for that reason.

Smaller budgets recover much of the score. On HotpotQA a $B{=}200$ run reaches
$0.717$ against the $B{=}500$ run's $0.727$, at a quarter of \gepa{}'s rollouts.
On Chia the $B{=}500$ run is what clears \gepa{}-light outright, but a smaller
run self-stops at $135$ rollouts and reaches $0.553$, within one standard error
of \gepa{}-light's $0.562$ and at fewer tokens.

\begin{table}[!ht]
\centering
\small
\setlength{\tabcolsep}{5pt}
\begin{tabular}{ll rrr}
\toprule
\textbf{Bench} & \textbf{Method} & \textbf{Rollouts} & \textbf{Tokens} & \textbf{Wall (s)} \\
\midrule
\multirow{3}{*}{Chia}
  & \rlmopt{} (light) & $135$ & $3.11$M & $1{,}317$ \\
  & \rlmopt{} (heavy) & $\mathbf{500}$ & $4.58$M & $1{,}357$ \\
  & \gepa{}-light     &$842$ & $3.81$M & $1{,}244$ \\
\midrule
\multirow{3}{*}{HotpotQA}
  & \rlmopt{} (light) & $\mathbf{200}$ & $\mathbf{1.78}$M & $\mathbf{717}$ \\
  & \rlmopt{} (heavy) & $500$ & $3.51$M & $1{,}673$ \\
  & \gepa{}-light     &$784$ & $2.53$M & $1{,}359$ \\
\midrule
\multirow{3}{*}{IFBench-25}
  & \rlmopt{} (light) & $200$ & $1.01$M & $1{,}163$ \\
  & \rlmopt{} (heavy) & $\mathbf{500}$ & $\mathbf{1.52}$M & $\mathbf{1{,}656}$ \\
  & \gepa{}-light     &$781$ & $1.78$M & $1{,}921$ \\
\midrule
\multirow{3}{*}{BFCL-mt}
  & \rlmopt{} (light) & $200$ & --- & $1{,}036$ \\
  & \rlmopt{} (heavy) & $\mathbf{475}$ & $5.2$M & $\mathbf{1{,}854}$ \\
  & \gepa{}-light     &$639$ & $7.05$M & $5{,}344$ \\
\bottomrule
\end{tabular}
\caption{Per-run compute (single seed), \rlmopt{} against \gepa{}-light. Bold
marks the cheapest \rlmopt{} run that still beats \gepa{}-light. The rollout
counts are downstream evaluations only; the text above gives the full
task-LM accounting, in which harness-side scoring dominates.}
\label{tab:budget}
\end{table}

\subsection{Optimized prompt sizes}
\label{sec:results:prompts}

\rlmopt{}'s prompts are smaller than \gepa{}'s while scoring higher, at
$27$--$79\%$ of \gepa{}-light's size on every benchmark
(Table~\ref{tab:promptsize}). The largest difference is on BFCL multi-turn,
where the winning instruction is $5{,}419$ characters against $19{,}818$. Appendix~\ref{app:prompts} provides the seeds and optimized prompts in detail.

\begin{table}[tb]
\centering
\small
\setlength{\tabcolsep}{7pt}
\begin{tabular}{l rr rr}
\toprule
\textbf{Benchmark} & \textbf{\gepa{}-light} & \textbf{\rlmopt{}}
& \textbf{$\Delta$} & \textbf{\% of \gepa{}} \\
\midrule
HotpotQA     & $5{,}043$  & $3{,}961$  & $-1{,}082$  & $79\%$ \\
IFBench-25   & $7{,}744$  & $3{,}073$  & $-4{,}671$  & $40\%$ \\
BFCL-mt      & $19{,}818$ & $5{,}419$  & $-14{,}399$ & $\mathbf{27\%}$ \\
Chia         & $24{,}556$ & $12{,}605$ & $-11{,}951$ & $51\%$ \\
\bottomrule
\end{tabular}
\caption{\textbf{Optimized-prompt size} (characters, seed $7$). Sizes are the
instruction portion; \rlmopt{} additionally injects worked-example demos.}
\label{tab:promptsize}
\end{table}

\section{Analysis}
\label{sec:analysis}

\subsection{Agent trajectories and prompt improvements}
\label{sec:analysis:trajectory}

Across runs, the most common first-action class is \emph{schema introspection}. Initial actions typically consist of \texttt{describe\_task} or \texttt{peek\_examples}, with the first \texttt{run\_candidate} call occurring only after several introspection steps. HotpotQA provides a representative example. The agent's trajectory converges on an \texttt{output-format-discipline} strategy: it inspects $3$--$5$ gold answer spans, identifies that the references favor short factual spans rather than complete sentences, and then commits a candidate with a new \texttt{OUTPUT FORMAT} section that explicitly enforces shortest-span extraction. Thus, the resulting prompt modification is grounded in observed reference outputs rather than in undirected candidate search.

For HotpotQA, the \rlmopt{}-optimized prompt improves test F1 from $0.700$ for the seed prompt to $0.727$ (seed $7$; Table~\ref{tab:headline}). We reproduce the full optimized prompt in Appendix~\ref{app:prompts}, and Figure~\ref{fig:trajectory} illustrates how the agent's tool trajectory translates into the resulting prompt structure.

\begin{figure}[!ht]
\centering
\small
\begin{tabular}{@{}c p{0.04\linewidth} p{0.20\linewidth} p{0.32\linewidth} p{0.28\linewidth}@{}}
\toprule
& \textbf{\#} & \textbf{Tool call} & \textbf{What the agent observed} & \textbf{Prompt section produced} \\
\midrule
\multirow{5}{*}{\rotatebox{90}{\textit{$\downarrow$ time}}}
& 1 & \texttt{describe\_task} & input/output schema: \texttt{question}, \texttt{context} $\to$ \texttt{answer:str} & \textit{(preamble: restate task)} \\
\addlinespace[2pt]
& 2 & \texttt{peek\_examples} \scriptsize $(n=3)$ & gold answers are \emph{short spans}: ``Princeton Rays'', ``no'', ``Tomas Arana'', never sentences & \texttt{\#\# OUTPUT FORMAT} \scriptsize\textit{(the key add)} \\
\addlinespace[2pt]
& 3 & \texttt{score\_explain} & scoring is word-level F1; partial credit for substrings; strict on extra punctuation & \texttt{\#\# STRICT FIELD FORMATTING} \\
\addlinespace[2pt]
& 4 & \texttt{scratchpad\_add} \scriptsize $\times 4$ & 4 facts promoted to rules: short-span rule; yes/no for comparisons; no ``The answer is''; 2 worked examples & \texttt{\#\# RULES}, \texttt{\#\# EXAMPLES}, \texttt{\#\# DOMAIN GUIDANCE} \\
\addlinespace[2pt]
& 5 & \texttt{commit\_prompt} & — & commits a strictly-better candidate \scriptsize\textit{(from introspection alone; $0$ \texttt{run\_candidate} calls)} \\
\bottomrule
\end{tabular}
\caption{\textbf{Optimization trajectory} on HotpotQA seed $7$, reconstructed
from the run's audit log. This is an illustrative (above-mean) ``surgical-mode''
trajectory where the agent self-stops \emph{without spending any
\texttt{run\_candidate} budget}: the four introspection tool calls
generate enough structure-relevant observations to commit a
strictly-better candidate against the seed. The trace-conditioned move
is step $2$: peeking $3$ gold examples reveals that answers are short
factual spans, which drives the \texttt{\#\# OUTPUT FORMAT} section of the
optimized prompt (Appendix~\ref{app:prompts}).}
\label{fig:trajectory}
\end{figure}

\subsection{When prompt optimization helps}
\label{sec:analysis:failure}
\label{sec:analysis:headroom}

Prompt optimization is typically evaluated on tasks where it produces gains, making it difficult to predict when optimization will be useful on a new task. Our results indicate that the key factor is the amount of prompt-accessible headroom left by the seed. We observe two distinct regimes: tasks where the task LM can exploit substantial remaining headroom, and tasks where the seed is already near the model's prompting ceiling.

\paragraph{Headroom regime.}
All four headline benchmarks fall in the first regime. Their seed prompts leave room for improvement that the task LM can exploit, with the largest gains occurring on the tasks with the most headroom. Chia starts at $0.435$ and improves by $+0.133$ to $0.568$, while the agentic BFCL task improves from $0.602$ to $0.686$ ($+0.084$). HotpotQA and IFBench-2025 begin closer to the task LM's prompting ceiling and consequently show smaller gains of $+0.027$ and $+0.030$, respectively (Table~\ref{tab:headline}). A synthetic diagnostic provides a controlled test of this mechanism: when the labels can be recovered from the training data, the optimizer discovers the relevant rule and generalizes it to held-out inputs. The same diagnostic also exposes an overfitting regime in which increasing the search budget reduces held-out accuracy (Appendix~\ref{app:headroom}).

\paragraph{Ceiling regime.}
When the seed is already near the task LM's prompting ceiling, additional optimization provides little opportunity for improvement. We observed this when evaluating capability-saturated tasks. On a mid-size model, classic IFEval-style instruction following achieved approximately $0.91$ from the seed, while the single-turn BFCL split reached approximately $0.77$. Neither \rlmopt{} nor \gepa{} improved these scores. In these cases, \rlmopt{}'s no-regression floor returned the seed prompt rather than accepting a noisy candidate with a lower score. We therefore treat these tasks as negative controls rather than headline results: they demonstrate that additional search does not create gains when prompt-accessible headroom is already exhausted.

\paragraph{Implications.}
The two regimes explain why optimization budgets can produce substantial gains on some tasks but little or no improvement on others. The relevant quantity is not the search budget itself, but the exploitable headroom remaining in the seed prompt. When a task is near the model's prompting ceiling, increasing the optimization budget is unlikely to help; improving the underlying task model is the more promising source of additional performance.

\paragraph{Self-stopping.}
The synthetic diagnostic also reveals a limitation of the current stopping policy. The agent's stopping point varies substantially: it used $27$ and $21$ rollouts in the two runs reported in Appendix~\ref{app:headroom}, and fewer on several other tasks. Increasing the budget cap therefore does not reliably increase the amount of search or improve held-out accuracy; in the larger-budget diagnostic run, test accuracy was lower. This variability suggests that the stopping decision remains an important source of run-to-run variation. A policy that explicitly considers remaining per-field headroom before terminating could make single runs more reliable (Section~\ref{sec:limitations}).

\section{Limitations}
\label{sec:limitations}

\paragraph{System-level attribution.}
\rlmopt{} differs from the baselines along several dimensions simultaneously, including its adaptive search policy, composite per-field scoring, no-regression floor, and optimizable demonstrations component. Our experiments therefore establish the performance of the complete system, but do not isolate the contribution of each component. A stronger attribution study would hold the harness, budget, task LM, and optimizer LM fixed while replacing the adaptive policy with a fixed search procedure. We leave this ablation to future work. This distinction is particularly relevant for BFCL multi-turn, where \rlmopt{} optimizes a demonstrations component that is not exposed by our \gepa{} wrapper. The resulting margin should therefore be interpreted as evidence for the combined system rather than for the adaptive search policy alone.

\paragraph{Stopping policy.}
The current stopping policy remains a source of variance. The three stopping conditions described in Section~\ref{sec:method:stop} are not well calibrated to the settings evaluated here. The per-field target of $0.85$ is unattainable for the composite scores observed in our benchmarks (Table~\ref{tab:headline}), while the $0.02$ improvement threshold is smaller than the estimated per-run standard error of $0.03$--$0.05$ (Section~\ref{sec:experiments}). The budget-based condition also did not determine termination in the trajectory shown in Section~\ref{sec:analysis:trajectory}. In practice, the budget cap is therefore the principal effective constraint. This contributes to substantial variation in stopping points across runs; in the synthetic diagnostic of Appendix~\ref{app:headroom}, increasing the budget cap even resulted in lower held-out accuracy. A better stopping policy would operate on quantities already available to the system, such as remaining per-field headroom and the statistical criterion used for commit decisions. We have not evaluated such a policy, so our reported sample-efficiency results reflect the current stopping behavior rather than an optimized termination strategy.

\paragraph{Evaluation scope and cost.}
Our evaluation covers four benchmarks and two task LMs: \texttt{gpt-4.1} for Chia and \texttt{gpt-4o-mini} for the remaining benchmarks. All four benchmarks were selected to leave exploitable headroom, so the results do not establish how the method behaves on tasks that are already near the task LM's prompting ceiling or on smaller open models. The observed margins over \gepa{} are also generally comparable to the per-run standard error (Section~\ref{sec:results:headline}); consequently, the evidence for an advantage comes from consistency across benchmarks and seeds rather than from large margins on individual runs.

Our compute comparison likewise concerns downstream candidate evaluations rather than total LM cost. Harness-side scoring accounts for a substantial fraction of task-LM calls (Section~\ref{sec:analysis:apicalls}), while the optimizer agent's reasoning uses a separate, stronger model whose cost is accounted for in token usage. Finally, the multi-component extension described in Section~\ref{sec:extension} remains sensitive to small validation sets. In an early probe of a mixed function-calling task with approximately ten validation and eight test examples, optimization improved validation performance on every seed ($+0.067$ on average) but did not improve held-out performance ($\Delta \approx -0.03$). This illustrates the risk of validation overfitting when the evaluation set is small and motivates larger validation sets for future extensions.

\section{Conclusion}
\label{sec:conclusion}

\rlmopt{} replaces a hand-coded outer prompt-search loop with an RLM agent that serves as the search policy, backed by a deterministic harness that enforces composite per-field scoring, Pareto selection, budget constraints, and a no-regression floor. Across four benchmarks spanning clinical extraction, multi-hop question answering, verifiable instruction following, and multi-turn agentic tool calling, \rlmopt{} achieves the best held-out score on every benchmark and outperforms \gepa{} on the four-task mean while using fewer downstream rollouts. Although the per-benchmark margins are generally comparable to the per-run standard error, the matched multi-seed comparison provides stronger evidence of reliability. Across all 11 benchmark--seed comparisons, \rlmopt{} never produces a prompt that underperforms its seed, while \gepa{} does so twice.

The results also clarify when prompt optimization can be expected to help. The available improvement is determined primarily by the headroom left by the seed relative to the task LM's prompting ceiling, rather than by the search budget alone. When a seed is already near that ceiling, additional optimization yields little benefit, and a stronger target model is the more effective source of improvement. When headroom remains, the relevant objective is therefore not simply to search more, but to reach the available improvement efficiently and reliably.

This perspective shifts the goal of prompt optimization from maximizing search to making effective use of the headroom that exists. A useful optimizer should improve weak seeds when prompt-level gains are available, avoid regressions when they are not, and do so without requiring excessive search. In this sense, knowing when optimization cannot help is an important part of knowing how to optimize.


\appendix
\clearpage
\section{The Optimizer's Skill Prompt}
\label{app:metaprompt}

The agent is conditioned by a fixed \emph{skill prompt}, the analogue of a
reflective optimizer's meta-prompt, which states the search discipline it should
follow and the structure every committed candidate must have. It is not modified
during optimization. We excerpt the load-bearing parts below verbatim, with
elisions marked \texttt{[...]}. The
excerpt is reproduced exactly as the agent receives it, so it refers to the
harness by its name in the implementation, \texttt{host}.

\begin{promptbox}[rlmopt]{\rlmopt{} optimizer skill (excerpt)}
{\scriptsize
\begin{verbatim}
You are a prompt optimizer. Your task is to maximize a deterministic scorer
by iteratively rewriting a `skill_instructions` prompt for a downstream
agent. Every candidate prompt you produce MUST follow the structure rules
below -- they are not optional.

# Core loop (every run -- do these in order)

These are the load-bearing rules; everything below elaborates them. If a
detailed section ever seems to conflict with this loop, follow the loop.

  1. ORIENT ONCE -- FIRST PASS ONLY. Call describe_task() +
     dataset_overview() + peek_examples() a single time each, then STOP --
     they are static. [...]
  2. INHERIT prior learning. Read describe_task()['known_rules'] (rules
     promoted from earlier runs of THIS task) AND scratchpad_read(...).
     Apply them; don't re-derive what a prior run proved.
  3. BASELINE, then DIAGNOSE from FAILURES, not random examples. Run the
     seed once, then write your candidate from ERROR-DRIVEN evidence:
     describe_failure_patterns (aggregates failures across ALL evaluated
     examples -- this is how you use the whole train set, not a 3-example
     skim) + peek_failures(n=6) (the concrete failing cases with the WHY
     diagnosis). Name ONE concrete failure mode, then edit to fix it.
  4. EDIT with intent, then SCORE representatively. Make ONE targeted
     change, then run_candidate(prompt) WITHOUT example_ids (the host's
     representative minibatch). NEVER score on your hand-picked failures --
     that subset anti-correlates with the real full-val selection.
  5. TRUST the verdict. Build only on REAL_GAIN. On WITHIN_NOISE do NOT
     commit or re-run to confirm -- attack a DIFFERENT failure mode.
     Bigger is NOT better.
  6. CAPTURE what worked. When a change earns a REAL_GAIN, record the
     durable, transferable rule via scratchpad_add(kind="rule", ...) -- it
     is promoted to the cross-run library so future runs start ahead.

# Structure rules (apply to every candidate you commit)

  **Bigger is NOT better -- test a simpler variant too.** A longer, more
  detailed prompt often scores WORSE than a concise one, especially for
  structured or short-output tasks and smaller target models: extra
  instructions dilute and distract. Do NOT assume adding detail helps. If
  your elaborated candidates are not beating the running best, test the
  OPPOSITE direction -- run_candidate a MINIMAL variant (close to the seed,
  or shorter than your current best). Keep whichever actually scores higher
  on full val; a good optimizer that can't improve should at least MATCH
  the seed, never only pile on text that loses to it.

  1. **Open with one crisp framing sentence** stating the task domain and
     the expected output shape. Do not start with "You are a helpful
     assistant" or generic preambles.

  2. **Section style depends on `style`.** [...]

  3. **Preserve every schema field name VERBATIM** from the task harness's
     `output_schema`. Do not rename, alias, translate, or invent new field
     names. Same for any enum values or fixed strings. [...]

[...]

# Anti-hallucination rules

  - example_ids you pass to run_candidate or peek_examples MUST come from
    describe_task()['train_example_ids'] or ['val_example_ids']. The host
    rejects unknown IDs.
  - score values come from run_candidate / score_explain. Do not invent
    them.
  - commit_prompt returns the truth about acceptance -- do not claim a
    commit succeeded if the host rejected it.

# Output

  - best_prompt:          the final text for the primary component
  - best_components_json: JSON {name: text} of ALL final component texts
  - audit_log:            a list of {iteration, action, rationale}
                          describing your search
\end{verbatim}
}
\end{promptbox}

Three parts of this text bear directly on results reported in the body. Step~3
is the diagnose-from-failures discipline whose effect is visible in the
trajectories of Section~\ref{sec:analysis:trajectory}; the ``bigger is not
better'' rule is why the optimized prompts stay short
(Section~\ref{sec:results:prompts}); and step~5 instructs the agent to treat a
within-noise result as uninformative, which complements the harness-side gate of
Section~\ref{sec:method:diagnosis} rather than replacing it. The skill is
advisory in all three cases: only the harness can refuse a commit.

\section{Optimized Prompts (Verbatim)}
\label{app:prompts}

This gallery reproduces, for every benchmark, the seed prompt (grey) and the
\rlmopt{}-optimized prompt (green) in full, so the structure the optimizer adds
is directly visible. The corresponding \gepa{}-light prompts run from $5{,}043$
to $24{,}556$ characters (Table~\ref{tab:promptsize}). Long lines are
wrapped to the box and text is ASCII-folded; the prompts are otherwise verbatim.

\subsection{Chia}
\begin{promptbox}[seed]{Chia --- seed (206 chars)}
{\scriptsize
\begin{verbatim}
Read the clinical trial eligibility criteria and extract the medical entities 
mentioned. For each entity type, list the exact text spans found in the criteria. 
Return empty lists for types with no mentions.
\end{verbatim}
}
\end{promptbox}
\begin{promptbox}[rlmopt]{Chia --- \rlmopt{} (12,605 chars)}
{\scriptsize
\begin{verbatim}
Extract clinical-trial eligibility entities from one criteria block as lists of 
exact text spans for six medical categories.

## Input
- You receive a single string field "criteria" containing one eligibility 
criterion or a short block of criteria text from a ClinicalTrials.gov protocol.

## Output Format
Return a Python-style JSON-compatible object with exactly these six fields, each a
 list of strings (0 or more items):
- conditions: list[str]
- drugs: list[str]
- procedures:
  - Extract diagnostic or therapeutic interventions performed on the patient, 
including:
    - surgeries and operations (e.g., 'vitrectomy surgery', 'submacular surgery', 
'surgical intervention').
    - tests and imaging (e.g., 'serological testing', 'biopsy', 'Urine ?-HCG 
pregnancy test').
    - treatments or courses of therapy (e.g., 'progesterone therapy', 
'treatment').
  - Separate procedures from conditions and measurements:
    - If a phrase combines a procedure with a condition (e.g., 'biopsy-proven 
LN'),
      extract 'biopsy' as a procedure and 'LN' as a condition.
    - Laboratory analytes or values (e.g., 'ALT', 'Serum creatinine', '2-hour 
C-peptide level')
      belong in measurements, not procedures; only the act of testing (e.g., 
'serological testing')
      is a procedure.
  - Do not include general eligibility language or follow-up phrases as procedures
    (e.g., 'follow-up visit', 'scheduled visit') unless they clearly denote a 
specific medical
    intervention or test.
  - When multiple similar procedures are listed together, extract each as a 
separate span
    if the gold-standard patterns favor separate items (e.g., 'vitrectomy 
surgery',
    'submacular surgery', 'surgical intervention').
- measurements: list[str]
- temporals: list[str]
- observations:
  - Capture ONLY clinical status, findings, or ongoing routines about the patient,
 such as:
    - symptoms or signs (e.g., 'pregnant', 'breastfeeding', 'acute intercurrent 
illness')
    - stable care patterns or routines (e.g., 'regular bowel care routine')
    - descriptive phrases that summarize current health state (e.g., 'good 
health').
  - Do NOT include legal, administrative, or cooperation requirements (e.g., 
'willing to participate',
    'able to comply with study procedures', 'provide written informed consent') as
 observations.
  - Do NOT move entities that belong to other fields into observations:
    - conditions remain in conditions (e.g., 'Impaired liver function', 'MI', 
'CVA').
    - procedures remain in procedures (e.g., 'serological testing', 'treated').
    - measurements remain in measurements (e.g., 'ALT', 'Serum creatinine').
  - Prefer under-extraction over over-extraction for observations: if a phrase is 
ambiguous between
    an administrative requirement and a clinical finding, leave it out of 
observations unless it
    clearly describes patient status.
  - Treat environmental or lifestyle exposure phrases that describe a patient's 
current situation
    as observations when they do not clearly denote a disease, for example:
    - 'altitude exposure', 'current heavy smoking', 'occupational exposure', 'high
 altitude exposure'.
  - In these cases, keep the disease or risk-condition in conditions (e.g., 
'respiratory disease',
    'cardiovascular disease') and put only the exposure/state itself in 
observations (e.g., 'altitude exposure').
  - Do not create observations entries for phrases that simply restate a condition
 name with modifiers;
    those should remain in the conditions list.
- conditions: Diagnoses, syndromes, or clinical states describing a disease or 
health condition (e.g., "acute coronary syndrome", "diabetes", "stroke").
- drugs: Names of medications or biologic agents, including abbreviations and 
combination phrases (e.g., "beta-blocker", "dual antiplatelet therapy").
- procedures: Surgical or interventional procedures, medical operations, or 
therapeutic interventions (e.g., "liver transplant", "angioplasty").
- measurements: Quantitative or score-based measures, thresholds, or lab values 
(e.g., "LDL-cholesterol", "CrCl", "Beck's Depression Inventory (BDI)").
- temporals: Words or phrases indicating time, duration, or ordering of events 
(e.g., "prior", "history", "after pretreatment", "within 6 months").
- observations: Non-diagnostic patient attributes, behaviors, or clinical findings
 that are not clearly procedures or measurements (e.g., "regular bowel care 
routine", "pregnant", "breastfeeding").

## Field-specific rules

- conditions:
  - Extract each disease or clinical condition as its own span, even if listed in 
a series.
  - Include both expanded and abbreviated forms when both appear (e.g., extract 
"Acute coronary syndrome" and "ACS" separately).
  - Prefer condition phrases (e.g., "bleeding") over generic outcome words like 
"death" when they appear together, but include "death" when explicitly listed 
among clinical outcomes.

- drugs:
  - Treat drug classes, specific drug names, and combination therapies as drug 
spans.
  - When a therapy phrase mixes drug and procedure concepts (e.g., "triple 
antiplatelet therapy"), extract it as drugs only if it clearly refers to 
medication combinations.
  - Do not duplicate abbreviations that are already part of the same phrase; 
extract each distinct token or phrase once.

- procedures:
  - Extract concrete interventions and operations (e.g., "Bioresorbable Vascular 
Scaffold implantation", "repeat revascularization", "liver transplant").
  - Do not label generic words like "treatment" alone as procedures when no 
specific procedure is given.
  - When a phrase combines therapy and procedure (e.g., "triple antiplatelet 
therapy", "dual antiplatelet therapy"), prefer classifying it as drugs unless 
explicitly described as a procedure.

- measurements:
  - Extract the name of the measurement itself, not the full inequality or numeric
 bound.
    - Example: from "CrCl < 30ml/min", extract "CrCl".
    - Example: from "Beck's Depression Inventory (BDI) =14", extract "Beck's 
Depression Inventory (BDI)".
    - Example: from "LDL-cholesterol> 1.8 mmol / l", extract "LDL-cholesterol".
  - If multiple numeric thresholds are given for the same measurement, still 
extract the single measurement name once.
  - Do not treat plain counts of events (e.g., "2 previous failed embryo 
transfers") as measurements unless a named scale or lab test is present.

- temporals:
  - Extract short time-related words or phrases that indicate timing, duration, or
 ordering of events.
    - Examples that should be extracted as temporals when they refer to past 
events:
      - "history" (when used as in "history of myocardial infarction").
      - "prior" and "previous" (when they indicate that a condition or procedure 
occurred in the past).
      - "after pretreatment".
      - Phrases like "within 6 months", "at least 4 weeks", "for at least 4 
weeks".
  - Do **not** extract purely numeric count phrases as temporals when they 
primarily indicate quantity rather than time.
    - Example: from ">= 2 previous failed embryo transfers", do **not** extract 
">= 2 previous failed embryo transfers" as a temporal.
    - Instead, treat "previous" as a temporal only if it can be extracted as a 
minimal standalone temporal word or short phrase.
  - Prefer minimal temporal spans that still convey the temporal meaning (e.g., 
"after pretreatment", "history", "prior", "previous").
  - Do not re-label measurement thresholds (e.g., "CrCl < 30ml/min") as temporals.

- observations:
  - Use observations for patient status, routines, and non-diagnostic findings 
that describe how the patient lives or functions, not for diseases or procedures.
    - Example: extract "regular bowel care routine".
    - Example: extract status phrases like "pregnant" and "breastfeeding" when 
they describe current patient status.
  - Do **not** extract purely legal, administrative, or cooperation-related 
phrases as observations:
    - Examples to skip: "No legal ability or legal ability is limited", "subjects 
unlikely to cooperate".
  - When a phrase mixes administrative and clinical content, extract only the 
clinical part if it can be isolated as a verbatim substring; otherwise skip.
  - Prefer to skip borderline administrative or compliance phrases rather than 
over-extract them as observations.
  - Do not re-label conditions or procedures as observations; leave them in their 
primary category only.

## General guidance

- Focus on high recall for conditions, drugs, procedures, measurements, and 
temporals while avoiding over-extraction for observations.
- Each output list may be empty or contain multiple spans; order does not matter 
for scoring.
- Ensure that every extracted span is present verbatim in the input criteria text.

# WORKED EXAMPLES (follow this input -> output mapping)

Example 1:
Input: {"criteria": "Adults older than 45 and children younger than 18 
years\nPlatelet count higher than 30x109/l at time of screening\nSuspicion of 
secondary ITP\nPositive family history for ITP\nPresence or history of autoimmune 
disease as judged by the investigator\nHepatosplenomegaly\nPresence or history of 
relevant hepatic disease as judged by the investigator\nPresence or history of 
thromboembolic disease as judged by the investigator\nPatients with 
splenectomy\nWomen who are pregnant or breast feeding\nIntention to become 
pregnant during the course of the study\nLack of safe double contraception (see 
7.1)\nAny vaccination 2 weeks prior start of the study\nDrugs with a known impact 
on the immune system or on platelet function must be recorded and an exclusion of 
the study should be discusse
Output: {"conditions": ["secondary ITP", "autoimmune disease", "hepatic disease", 
"thromboembolic disease", "alcohol abuse", "drug abuse", "Hypersensitivity", 
"Hepatosplenomegaly"], "drugs": ["romiplostim", "eltrombopag"], "procedures": 
["splenectomy", "vaccination"], "measurements": ["Platelet count"], "temporals": 
["at time of screening", "2 weeks prior start of the study"], "observations": 
["family history for ITP"]}

Example 2:
Input: {"criteria": "Clinically significant systemic disease (such as diabetes, 
metabolic syndrome, immunological diseases, diagnosed thrombophilia, porphyria, or
 any other medical condition requiring the use of low-molecular weight heparin 
therapy)\nPolycystic ovary syndrome (PCOS) according to Rotterdam Consensus 
Criteria (European Society of Human Reproduction and Embryology [ESHRE]/American 
Society for Reproductive Medicine [ASRM], 2003)\nPoor ovarian response (POR) 
according to the European Society of Human Reproduction and Embryology (ESHRE) 
Criteria\nRIF (repeated implantation failure), defined as greater than or equals 
to (>=) 2 previous failed embryo transfers\nEndometriosis III-IV stage or 
adenomyosis\nClinically significant findings on exam or ultrasound, such as 
salpingitis, hydrosalp
Output: {"conditions": ["systemic disease", "diabetes", "metabolic syndrome", 
"immunological diseases", "diagnosed thrombophilia", "porphyria", "medical 
condition", "Polycystic ovary syndrome (PCOS)", "Poor ovarian response (POR)", 
"RIF (repeated implantation failure)", "Endometriosis", "adenomyosis", 
"salpingitis", "hydrosalpynx", "ovarian cysts", "hypersensitivity"], "drugs": 
["low-molecular weight heparin", "vaginal progesterone", "excipients", "components
 of the solution"], "procedures": ["exam", "ultrasound"], "measurements": 
["previous failed embryo transfers"], "temporals": [], "observations": 
["findings"]}

Example 3:
Input: {"criteria": "Patients must have histologic proof of a malignancy suitable 
for radiation therapy. \nPatients must have received prior external beam radiation
 therapy to the region proposed for HDR brachytherapy treatment; evaluation of 
doses previously delivered to spinal cord/cauda equine, pelvis, and other critical
 structures (bowel, kidneys, rectum) will be taken into consideration. \nIf repeat
 irradiation would exceed any normal tissue constraint set by MSKCC Radiation 
Oncology Department dose constraint criteria, the patient will potentially be 
eligible. \nIf the total prior radiation dose to the cord or pelvis exceeds 100 Gy
 BED equivalent, the patient will be potentially eligible, where a total of 100 
BED Gy equivalent is determined by the biological equivalent dose (BED) calculatio
Output: {"conditions": ["malignancy"], "drugs": [], "procedures": ["radiation 
therapy", "histologic", "external beam radiation therapy", "HDR brachytherapy", 
"repeat irradiation"], "measurements": ["MSKCC Radiation Oncology Department dose 
constraint criteria", "KPS"], "temporals": ["prior"], "observations": []}
\end{verbatim}
}
\end{promptbox}

\subsection{HotpotQA}
\begin{promptbox}[seed]{HotpotQA --- seed (139 chars)}
{\scriptsize
\begin{verbatim}
Answer the multi-hop question using only the supporting paragraphs. Produce a 
short factual answer (a name, date, number, or short phrase).
\end{verbatim}
}
\end{promptbox}
\begin{promptbox}[rlmopt]{HotpotQA --- \rlmopt{} (3,961 chars)}
{\scriptsize
\begin{verbatim}
Answer the multi-hop question using only the information in the provided 
supporting paragraphs.

You must:
- Carefully read all paragraphs in the context, not just the first one.
- Identify the specific entities, dates, numbers, or facts mentioned in the 
question.
- Find the necessary pieces of evidence in two or more different paragraphs and 
mentally chain them together to reach the answer.
- Ignore any knowledge that is not explicitly stated in the context; do not rely 
on outside or prior knowledge.
- When the question is a comparison or verification (e.g., asking if a statement 
is true), use the context to decide and answer with a single word: "yes" or "no".
- When the question asks for a person, place, organization, date, number, title, 
or short phrase, copy the answer as a short span exactly as written in the 
context.

Input format:
- question: a multi-hop question that typically requires combining facts from at 
least two different paragraphs.
- context: several Wikipedia-style paragraphs; they contain all facts needed to 
answer the question.

Output format:
- Return only the short factual answer span, with no explanation, no extra words, 
and no leading phrases like "the answer is".
- Do not output full sentences.
- If multiple valid surface forms exist, choose the form that appears most 
directly and clearly in the context.

# WORKED EXAMPLES (follow this input -> output mapping)

Example 1:
Input: {"question": "Which 2008 French film was the third of a four-film 
adaptation?", "context": "## 2008 French Open - Boys' Singles\nThe 2008 French 
Open - Boys' Singles tournament was an event during the 2008 French Open tennis 
tournament.  Vladimir Ignatic was the defending champion, but did not compete in 
the Juniors in this year.\n\n## Asterix at the Olympic Games (film)\nAsterix at 
the Olympic Games (French: \"Ast?rix aux Jeux Olympiques\" ) is a 2008 French 
fantasy comedy film directed by Fr?d?ric Forestier and Thomas Langmann, and 
written by Langmann, Alexandre Charlot, and Frank Magnier, based on characters 
from Ren? Goscinny and Albert Uderzo's Ast?rix comic series.  It was filmed 
primarily in Spain over the course of the year 2006.\n\n## The Beautiful 
Person\nThe Beautiful Person (Fr
Output: "Asterix at the Olympic Games"

Example 2:
Input: {"question": "Who rode with the cosmonaut who commanded the historic 
Voskhod 2 mission?", "context": "## Pete Conrad\nCharles \"Pete\" Conrad Jr. (June
 2, 1930?- July 8, 1999), (Captain, USN), was an American NASA astronaut, naval 
officer and aviator, test pilot, and during the Apollo 12 mission became the third
 man to walk on the Moon.  He set an eight-day space endurance record along with 
his Command Pilot Gordon Cooper on the Gemini 5 mission, and commanded the Gemini 
11 mission.  After Apollo, he commanded the Skylab 2 mission (the first manned 
one), on which he and his crewmates repaired significant launch damage to the 
Skylab space station.  For this, President Jimmy Carter awarded him the 
Congressional Space Medal of Honor in 1978.\n\n## Alexey Leonov\nAlexey 
Arkhipovich Leonov (Rus
Output: "Alexey Leonov"

Example 3:
Input: {"question": "What was the occupation of the person whose first serious 
lover was Richard Chanlaire?", "context": "## Stripper\nA stripper or exotic 
dancer is a person whose occupation involves performing striptease in a public 
adult entertainment venue such as a strip club.  At times, a stripper may be hired
 to perform at a bachelor party or other private event.\n\n## Marshman\nThe name 
Marshman is a family, or surname which originated in England and either refers to 
an occupation - namely a person whose job it was to work the marshes or it is 
derived from their residency possibly of Marsham in Norfolk, or in Mersham in 
Kent.  There is a strong settlement of the Marshman family in Wiltshire, 
especially near Dilton Marsh.\n\n## Roughneck\nRoughneck is a term for a person 
whose occupation i
Output: "composer and pianist"
\end{verbatim}
}
\end{promptbox}

\subsection{IFBench-2025}
\begin{promptbox}[seed]{IFBench-2025 --- seed (143 chars)}
{\scriptsize
\begin{verbatim}
Follow the user's instruction and satisfy every explicit constraint it states 
(counts, formats, keywords, positions). Output only the response.
\end{verbatim}
}
\end{promptbox}
\begin{promptbox}[rlmopt]{IFBench-2025 --- \rlmopt{} (3,073 chars)}
{\scriptsize
\begin{verbatim}
Generate exactly the response requested by the user while strictly satisfying 
every explicit constraint in the instruction.

You must:
1. Identify explicit constraints:
   - Read the instruction carefully and list every explicit constraint on counts, 
formats, exact phrases, keywords, ordering, and positions.
   - Treat requirements such as "output nothing", "return None", "leave blank", 
"do not respond", "respond only with X", or similar phrases as hard constraints on
 what you may output.

2. Handle empty or placeholder responses:
   - If the instruction requires no meaningful content (for example, "output 
nothing", "leave the response blank", "do not answer"), output an empty response 
or the exact placeholder specified (such as the single token "None"), and nothing 
else.
   - Do NOT add explanations, descriptions, or any extra characters around an 
empty or placeholder response.
   - If the instruction says to "respond only with" a specific string, output that
 string verbatim and do not add any other text.

3. Plan the response structure:
   - Decide the structure of your response so that all count constraints are 
satisfied exactly (no more, no fewer items, words, characters, sentences, or lines
 than requested).
   - Align required positions ("start with ...", "end with ...", "the nth item 
must be ...") with the ordering of the content you will produce.

4. Match formats and keywords exactly:
   - Follow any specified format precisely, including symbols, spacing, digit 
patterns, letter case, punctuation, and template structure.
   - Include every required keyword or phrase verbatim and exclude every forbidden
 keyword or phrase, respecting case sensitivity when specified.

5. Avoid non-response content:
   - When the instruction asks for a constructed string, list, pattern, label, or 
sequence, output that content directly, without explanations or topic summaries.
   - Do NOT define concepts, explain laws, or provide background information 
unless the instruction explicitly asks for an explanation.

6. Resolve apparent conflicts conservatively:
   - If constraints seem to conflict, obey the most explicit, machine-checkable 
constraints first (counts, exact strings, formats, positions).
   - Do not relax or ignore any stated constraint; choose the interpretation that 
best satisfies all explicit requirements without adding new assumptions.

7. Verify before responding:
   - Before you output anything, mentally check that your response satisfies all 
stated count, format, keyword, ordering, positional, and "no explanation" 
constraints.
   - If any constraint is not satisfied, adjust the response in your head and only
 then output it.

Output rules:
- Output only the final response content that the instruction asks for.
- Do NOT explain your reasoning, restate the constraints, or add meta-commentary.
- Do NOT add bullet points, section headers, labels, or other markers unless the 
user explicitly requests them.
- Do NOT add quotes, brackets, or extra text around the response unless the 
instruction explicitly requires that formatting.
\end{verbatim}
}
\end{promptbox}

\subsection{BFCL multi-turn}
\begin{promptbox}[seed]{BFCL multi-turn --- seed (215 chars)}
{\scriptsize
\begin{verbatim}
You are given a user request and a list of available functions (with JSON argument
 schemas). Decide which function call(s) satisfy the request and output them. If 
no available function applies, output an empty list.
\end{verbatim}
}
\end{promptbox}
\begin{promptbox}[rlmopt]{BFCL multi-turn --- \rlmopt{} (5,419 chars)}
{\scriptsize
\begin{verbatim}
Plan the complete sequence of tool calls for a multi-turn conversation, emitting 
exactly one JSON array named `tool_calls`.

## Task
- Read the entire conversation from start to finish, respecting turn order and 
conversational state.
- Identify every explicit and implicit user request that requires using the 
provided tools.
- Decompose each request into the minimal sequence of tool calls needed to satisfy
 it fully, in the order they should be executed.

## Input
- "conversation": the full multi-turn dialogue, with user and assistant turns.
- "functions": a JSON list describing each available tool:
  - Each function has a "name", "description", and a JSON schema for its 
"arguments".

You must:
- Use ONLY the tools whose "name" appears in the `functions` JSON; never invent 
new function names.
- For each function, construct "arguments" that exactly match its schema (field 
names, types, and required/optional status).
- Derive argument values strictly from the conversation and tool descriptions; do 
not guess values not supported by the context.
- Respect temporal and dependency constraints: resolve prerequisites first (e.g., 
look up an ID before using it), then perform dependent actions.

## Output Format
Emit a single JSON array assigned to `tool_calls`. The top-level output must be 
ONLY this array, for example:
[
  { "name": "function_name", "arguments": { "arg1": "value1", "arg2": 2 } },
  { "name": "another_function", "arguments": { "flag": true } }
]

Formatting rules:
- Do not wrap the array in any additional object or fields.
- Do not include comments, natural-language explanations, or reasoning outside the
 JSON.
- Each element in the array must be a JSON object with exactly two keys:
  - "name": the function name string, exactly as given in `functions`.
  - "arguments": a JSON object whose keys and value types follow that function's 
argument schema.


## Rules

- When choosing among similar tools, always:
  - Base the choice strictly on each function's "description" field and arguments 
schema from the provided functions list.
  - Select the tool whose documented purpose most directly satisfies the user's 
request; do not use a more generic or adjacent tool when a specific one exists.
- Arguments must match the tool's JSON schema exactly:
  - Include all required arguments with appropriate types and units.
  - Do not add extra arguments that are not defined in the schema.
  - Use the exact field names and value formats shown in the tool schema and 
conversation (e.g., correct symbol strings, option names, booleans true/false).
- When the gold sequences show repeated or mirrored calls (e.g., calling a 
function twice with different arguments to compare or update state), reflect that 
pattern:
  - Emit multiple calls to the same tool when needed, rather than collapsing them 
into one approximate call.
- Treat each tool_call as explicitly linked to a user request or a necessary 
prerequisite for it:
  - Avoid speculative or redundant calls that are not supported by the 
conversation or tool descriptions.


- When working with folders and files:
  - Use navigation tools like cd(folder) to move into the correct directory BEFORE
 invoking tools that operate on files there.
  - Use listing/inspection tools like ls() or similar to confirm file presence 
where appropriate.
  - Use content-reading tools like cat(file_name) BEFORE running tools that depend
 on that file's contents.
- When performing actions that require authentication or identity:
  - Call the appropriate login/authentication tool (e.g., message_login(user_id) 
or equivalent) BEFORE sending messages or performing actions that assume a 
logged-in state.
  - Do not assume you are logged in unless a prior tool_call in the plan logs in 
during this conversation.
- For multi-step goals:
  - Plan each goal as a sequence of tool_calls where earlier calls establish 
prerequisites (navigation, selection, reading, authentication) and later calls 
perform the requested action.
  - Include intermediate "check" or "inspect" calls (such as reading current state
 or configuration) when the expected gold sequences show them; do not skip them to
 shorten the plan.

- tool_calls: 
  - Include every required tool call needed to satisfy the user's requests; do not
 omit necessary steps.
  - Avoid spurious calls: only include tools that the user's request or the tool 
descriptions clearly require.
  - Preserve and reuse state across calls when appropriate (e.g., watchlists, 
created objects) instead of recomputing or overwriting without being asked.
  - Use the minimal sufficient sequence of calls; do not add exploratory or 
redundant calls.
  - When the conversation includes multiple independent tasks, include calls for 
each task in the single `tool_calls` array, ordered in the sequence they should be
 performed.
  - Ensure argument values are precise and consistent with the request (e.g., 
correct quantities, symbols, IDs), and avoid changing a value unless the user 
explicitly requests it.

General guidance:
- Prefer direct calls that accomplish what the user asked, rather than indirect or
 unnecessary chains.
- If the user asks to both perform an action and then report results, include 
tools for the action first and then tools to retrieve or expose the result.
- If information required for a tool argument is missing from the conversation and
 tool descriptions, omit that tool call rather than guessing.
\end{verbatim}
}
\end{promptbox}

\section{Synthetic Headroom Diagnostic}
\label{app:headroom}

To isolate the mechanism behind the headroom regime
(Section~\ref{sec:analysis:headroom}) in a controlled setting, we ran a small
synthetic task on which the correct answer is recoverable only from the training
data. We report it here rather than in the main results because the seed prompt scores
zero by design, so the before/after delta would overstate the gain one should
expect from a competently written seed. The diagnostic is intended to
demonstrate a mechanism, not to certify an effect size.

\paragraph{Task.}
The task maps a customer-support message to a label. Each of eight semantic
intents (for example \texttt{refund\_request} or \texttt{account\_locked}) is
assigned an \emph{arbitrary} two-character code (\texttt{K9}, \texttt{V3},
\dots) that carries no relationship to the message text, and each intent is
expressed through eight natural phrasings, for $64$ records scored by exact match
on the code. Because the codes are arbitrary, the intent-to-code table cannot be
guessed from the surface text; it is present only in the training pairs. The seed
prompt (\emph{``Read the customer message and output the correct 2-character
code. Output only the code.''}) therefore scores essentially zero, not because
the task is hard but because the mapping has been withheld from the prompt.

\paragraph{What the optimizer does.}
The optimizer reads the training pairs, reconstructs the intent-to-code table, and
writes it into the prompt as a lookup. Because the underlying language model
already generalizes surface phrasing to intent, the recovered table transfers to
held-out phrasings of the same intents. In a committed run the optimized prompt reached $0.769$ on the held-out test split,
with validation at $0.867$, using $27$ rollouts. The value of the diagnostic is
that it confirms the optimizer can move information from the data into the prompt
and have it generalize, rather than merely memorizing the training rows.

\paragraph{Overfitting under a larger budget.}
Additional budget did not help monotonically. A separate run at a larger budget
reached only $0.385$ on test while overfitting validation to $0.800$, and the
agent's self-stop point varied between the two runs ($27$ and $21$ rollouts). The
practical reading is that exploitable headroom, not raw budget, governs whether
optimization helps, and that a larger budget cap can degrade held-out accuracy
when the agent over-searches a small validation set
(Section~\ref{sec:limitations}).

\section{Optimizer Tool Interface}
\label{app:tools}

This appendix documents the behavior of the tool interface summarized in
Table~\ref{tab:tools} (Section~\ref{sec:method:tools}). The agent never sees the
raw task-LM API; every action it takes is a call to one of these tools.

\paragraph{Introspection.}
\texttt{describe\_task} returns the task schema, the output fields, and the
metric assigned to each field. \texttt{dataset\_overview} reports split sizes and
aggregate statistics. \texttt{peek\_examples} and \texttt{view\_example} return
training and validation records with gold outputs. \texttt{query\_examples}
retrieves records matching a filter. \texttt{score\_explain} runs the scoring
procedure on a hypothetical prediction and returns the resulting breakdown,
which lets the agent learn how a metric behaves without consuming budget.
\texttt{list\_metrics} and \texttt{list\_components} enumerate the metric catalog
and the optimizable components of the current candidate.
\texttt{read\_additional\_instructions} surfaces optional operator guidance.

\paragraph{Failure analysis.}
These tools form a discover, narrow, and inspect ladder over prior rollouts.
\texttt{describe\_failure\_patterns} aggregates recurring error modes across
evaluated examples, \texttt{search\_traces} locates rollouts matching a query,
\texttt{peek\_failures} returns the lowest-scoring examples, and
\texttt{read\_trace} returns a full rollout record.

\paragraph{Synthesis.}
\texttt{synthesize\_failures} performs sub-LM root-cause analysis over a set of
failures and returns a compressed summary, which keeps the top-level agent's
context small. \texttt{synthesize\_candidate} does the same for the rollouts of a
single candidate, returning a shared failure mode and a suggested rule rather
than prompt text. \texttt{call\_subagent} issues a nested recursive call for
hierarchical decomposition. Of this group only \texttt{merge\_candidates}
returns a prompt: it combines two existing candidates into one. Every other
candidate the agent evaluates or commits is text the agent wrote itself
(Section~\ref{sec:method:search}).

\paragraph{Evaluation and state.}
\texttt{run\_candidate} is the only tool that consumes evaluation budget: it
scores a candidate on $n$ validation examples and returns the structured
feedback of Section~\ref{sec:method:feedback}. \texttt{commit\_prompt} submits a
candidate to the harness, which enforces the length and rationale guards, requires
a genuine composite improvement, and blocks any commit that regresses an
individual field past the floor (Appendix~\ref{app:harness}). The agent may choose
which parent to extend from the per-example Pareto frontier
(\texttt{pareto\_frontier\_status}) and may query \texttt{best\_so\_far} and
\texttt{remaining\_budget}, but it cannot promote a candidate itself.
\texttt{scratchpad\_add} and \texttt{scratchpad\_read} maintain persistent typed
notes (\texttt{fact}, \texttt{hypothesis}, \texttt{rule}, \texttt{warning},
\texttt{decision}) across turns.

\paragraph{Sealing the held-out split.}
Test isolation is enforced at the tool boundary rather than by instruction.
\texttt{dataset\_overview} reports the test count but withholds test
identifiers, and \texttt{query\_examples} and \texttt{view\_example} refuse
test-split references outright. No prompting of the agent can therefore surface
test gold for encoding into a candidate.

\section{Harness Selection Rules}
\label{app:harness}

This appendix specifies the candidate eligibility, final selection, polish,
and diagnose-gate rules used by the harness. These rules are deterministic and
apply independently of the RLM agent's search decisions.

\paragraph{Candidate eligibility.}
A candidate $p$ is eligible for commit only if it satisfies the active
per-field regression constraints, improves the composite score over the
running best, and passes the length and rationale guards applied by
\texttt{commit\_prompt}. When the per-field regression guard is active, the
candidate must satisfy

\begin{equation}
s_f(p) \geq b_f - f_{\mathrm{floor}}
\end{equation}

for every field $f$, where $b_f$ is the current best score and
$f_{\mathrm{floor}}$ is the configured field-floor tolerance. The guard is
enabled automatically on datasets of at most $20$ records and can be enabled
explicitly on larger ones; it is enabled for every run reported in this paper,
whose datasets are all larger than that threshold.

Candidate improvements are additionally subject to the significance gate
described in Section~\ref{sec:method:harness}. The paired improvement over the
running best must exceed $1.65$ standard errors. Candidates that do not satisfy
the eligibility conditions are not committed and do not become the running
best.

\paragraph{Final selection.}
After the agent-controlled search terminates, the harness constructs a final
candidate set containing the seed prompt, every committed candidate, the
agent's claimed best candidate, and the polish variants described below. The
harness computes per-field validation scores for every candidate, constructs
the Pareto frontier, and selects the frontier candidate with the highest
composite score.

The agent's claimed best is therefore not accepted merely because the agent
identifies it as best. It competes with the seed, committed candidates, and
polish variants under the final selection procedure.

\paragraph{Polish stage.}
Before final selection, the harness generates up to five competing variants
from its own running-best candidate rather than from the candidate claimed by
the agent.

The first variant is a deterministic structural rewrite that re-attaches the
seed's opening task statement to the accumulated rules. This is intended to
recover task framing that may have become diluted through incremental edits.

The next three variants append $3$, $8$, and $15$ gold worked examples from the
training split. The number of demonstrations is therefore selected by
validation rather than fixed in advance.

The final variant is a sub-LM rewrite of the running-best candidate. It is
discarded if its returned prompt is less than half the length of the source
prompt, which guards against rewrites that remove accumulated instructions.

All surviving polish variants are evaluated on validation data and participate
in the same final Pareto-frontier selection as the candidates generated during
the agent-controlled search. Polish evaluations are part of the deterministic
finalization stage and do not consume the agent's search budget $B$.

\paragraph{Diagnose gate.}
The harness maintains a counter of consecutive evaluated candidates whose
scores fall within a predefined noise band around the running best. The noise
band is defined using a symmetric $1.5$ standard-error interval around the
running-best score.

After two consecutive near-best evaluations, \texttt{run\_candidate} appends
an advisory warning to its result. After three consecutive near-best
evaluations without an intervening diagnosis, the next
\texttt{run\_candidate} call returns an error, consumes no evaluation budget,
and instructs the agent to invoke \texttt{synthesize\_failures} on its weakest
field.

A candidate producing a sufficiently large score change resets the counter.
A successful call to \texttt{synthesize\_failures} also resets the counter.
Re-reading existing failures through
\texttt{describe\_failure\_patterns} or \texttt{peek\_failures} does not count
as diagnosis, since these calls do not synthesize a new failure analysis.

The refusal mechanism is bounded. After two consecutive refusals, one
candidate evaluation is admitted regardless of the counter state, after which
the gate is re-armed. This prevents a non-compliant agent from deadlocking the
search.

The gate is disabled when no sub-LM is configured, since
\texttt{synthesize\_failures} requires a sub-LM. The gate limits repeated
evaluation of statistically indistinguishable candidates but does not prescribe
which hypothesis the agent should test after diagnosis.
\section{Reproducibility}
\label{app:repro}
\sloppy

\subsection{Environment}

\begin{itemize}[noitemsep,topsep=2pt]
  \item Python $\geq 3.11$, dependencies pinned in
    \texttt{uv.lock} (tracked under version control)
  \item DSPy $\geq 3.2$
  \item Deno $\geq 1.40$ (the agent's tool layer runs in a sandboxed
    REPL (a Deno$+$Pyodide sandbox); without Deno the optimizer silently
    degrades to template-only mode, so the SDK constructor now hard-fails
    when Deno is absent on \texttt{PATH})
  \item Models: \texttt{gpt-4.1} (task LM for Chia) and \texttt{gpt-4o-mini}
    (task LM for the other benchmarks), with \texttt{gpt-5.1} as the
    optimizer/reflection LM, all at $\texttt{temperature}=0$,
    $\texttt{cache}=\texttt{False}$, per-call timeout $90$--$300$\,s. Base model
    names are pinned in the run configs; the exact API snapshots are those served
    by the deployment gateway at run time
\end{itemize}

\subsection{Default hyperparameters}

Table~\ref{tab:hyperparams} lists the library defaults. The head-to-head runs
of Section~\ref{sec:results} override \texttt{budget\_calls} as described in
Section~\ref{sec:experiments}; every other value is used as shown.

\begin{table}[!ht]
\centering
\small
\begin{tabular}{@{}l l p{0.46\linewidth}@{}}
\toprule
\textbf{Parameter} & \textbf{Default} & \textbf{Notes} \\
\midrule
\texttt{budget\_calls} & $30$ & Downstream rollouts the agent can spend; head-to-heads use the \texttt{heavy} preset ($B{=}500$) \\
\texttt{optimizer\_max\_iterations} & $40$ & Hard cap on REPL turns \\
\texttt{seed} & $7$ & Affects split shuffle + resample RNG \\
\texttt{commit\_policy} & \texttt{no\_field\_regression} & Reject any commit that drops a field past \texttt{field\_floor} \\
\texttt{field\_floor} & $0.05$ & Max allowed per-field drop under \texttt{no\_field\_regression} \\
\texttt{style} & \texttt{thorough} & Section-structured rewrites; \texttt{lean} = surgical edits for strong seeds / tiny data \\
\texttt{self\_stop\_floor\_pct} & $0.80$ & Min fraction of $B$ consumed before self-stop ($0.25$ in \texttt{lean}) \\
\texttt{polish\_variants} & \texttt{True} & Run the POLISH-phase variant competition (Section~\ref{sec:method}) \\
\texttt{select\_significance\_k} & $1.65$ & Std-error multiple a commit must clear (one-sided $95\%$) \\
\texttt{use\_skill\_library} & \texttt{True} & Cross-run learned rules; \textbf{disabled} for the \gepa{} head-to-heads \\
\bottomrule
\end{tabular}
\caption{Run-time configuration for \rlmopt{}. Values shown are the
library defaults; the benchmark head-to-heads override \texttt{budget\_calls}
to the \texttt{heavy} preset ($B{=}500$) and disable
\texttt{use\_skill\_library} for a fair comparison against \gepa{}.
They also set \texttt{commit\_policy}$=$\texttt{no\_field\_regression}
explicitly, so the per-field floor is active on every benchmark run regardless of
the $\leq\!20$-record auto-enable described in Section~\ref{sec:method:harness}.
Every value is exposed on \texttt{RLMOptConfig}.}
\label{tab:hyperparams}
\end{table}

\subsection{Seed and evaluation protocol}

A single random seed (we use $7$) propagates to the dataset shuffle that
produces the train, validation, and held-out test splits, and to any resampling
used during selection. Within each benchmark both compared methods share that
seed, the same splits, and the same task LM, so the held-out test set is
identical across methods and each row is a matched, example-for-example
comparison. The headline comparison is reported at this fixed seed, and
Section~\ref{sec:results:seeds} repeats it at further seeds to test whether the
ordering holds.

\medskip\noindent\textbf{Splits and fields.} Each benchmark is split by the
shared seed into train, validation, and held-out test partitions. The held-out
test sizes are $100$ examples (Chia, HotpotQA) and $50$ (IFBench-2025, BFCL
multi-turn); the per-benchmark train and validation sizes are fixed per
benchmark and held constant across methods and seeds. Chia is scored on six entity fields---conditions, drugs,
procedures, measurements, temporals, and observations---weighted equally in the
composite.

\subsection{Record schema}

Every benchmark run writes a \texttt{record.json} under
\texttt{experiments/gepa\_benchmark/results/} at the sub-path
\texttt{<method>/<bench>/seed<N>\_<model>/record.json}, with the following keys
(excerpt):

\begin{verbatim}
{
  "method": "rlm_opt_teacher" | "gepa_auto_light" | "baseline",
  "bench": "chia" | "hotpotqa" | "ifbench2025" | "bfcl_multiturn",
  "seed": 7,                   // single-seed protocol (Section 5.1)
  "task_lm": "gpt-4.1" | "gpt-4o-mini",  // gpt-4.1 for Chia, gpt-4o-mini otherwise
  "rollouts_used": int,        // for RLMOpt: downstream evals only
  "wall_clock_s": float,
  "val_score": float, "test_score": float,
  "per_example_test": {example_id: float, ...},  // used by aggregator
  "cost_usd_estimate": float,
  "extra": {
    "api_calls": int,           // from len(lm.history); post-patch only
    "prompt_tokens": int,
    "completion_tokens": int,
    "total_tokens": int,
    "cached_tokens": int,       // Azure auto-prefix-cache hits
    "cache_hit_rate": float
  }
}
\end{verbatim}

Every number in the paper's tables and CIs traces to these files.

\end{document}